\pdfoutput=1

\documentclass[11pt]{article}
\usepackage[table,dvipsnames]{xcolor}
\usepackage{algorithm2e}
\usepackage{booktabs}
\usepackage[]{acl}

\usepackage{url}

\usepackage{times}
\usepackage{latexsym}
\usepackage[T1]{fontenc}

\usepackage[utf8]{inputenc}

\usepackage{microtype}

\usepackage{inconsolata}

\usepackage{graphicx}               
\usepackage{tabularx}               
\newcolumntype{C}{>{\centering\arraybackslash}X}
\usepackage{multirow}               
\usepackage{diagbox}                
\usepackage{hhline}                 
\usepackage{color}                  
\usepackage{amsmath,amsthm}                
\usepackage{bm}
\usepackage{amssymb}                
\usepackage{mathtools}              
\usepackage{enumitem}               
\usepackage{booktabs}
\usepackage{subcaption}
\usepackage{bbm}

\makeatletter
\newcommand*\bigcdot{\mathpalette\bigcdot@{.5}}
\newcommand*\bigcdot@[2]{\mathbin{\vcenter{\hbox{\scalebox{#2}{$\m@th#1\bullet$}}}}}
\makeatother

\usepackage{xparse}
\NewDocumentCommand{\heng}{ mO{} }{\textcolor{OrangeRed}{\textsuperscript{\textit{Heng}}\textsf{\textbf{\small[#1]}}}}
\NewDocumentCommand{\qianying}{ mO{} }{\textcolor{CadetBlue}{\textsuperscript{\textit{Qianying}}\textsf{\textbf{\small[#1]}}}}
\NewDocumentCommand{\fei}{ mO{} }{\textcolor{Blue}{\textsuperscript{\textit{Fei}}\textsf{\textbf{\small[#1]}}}}
\NewDocumentCommand{\lingfei}{ mO{} }{\textcolor{Cyan}{\textsuperscript{\textit{Lingfei}}\textsf{\textbf{\small[#1]}}}}

\NewDocumentCommand{\aysa}{ mO{} }{\textcolor{Purple}{\textsuperscript{\textit{Aysa}}\textsf{\textbf{\small[#1]}}}}

 \usepackage{tabularray}
\usepackage{lipsum}  
\usepackage{longtable}
\usepackage{subcaption}
\usepackage{adjustbox}
\usepackage{tikz}
\usepackage{listings}
\usepackage{xcolor,colortbl}

\colorlet{punct}{red!60!black}
\definecolor{background}{HTML}{EEEEEE}
\definecolor{delim}{RGB}{20,105,176}
\definecolor{mycolor}{RGB}{173,216,230}

\colorlet{numb}{magenta!60!black}

\lstdefinelanguage{json}{
    basicstyle=\scriptsize\ttfamily,
    numbers=left,
    numberstyle=\scriptsize,
    stepnumber=1,
    numbersep=8pt,
    showstringspaces=false,
    breaklines=true,
    frame=lines,
    backgroundcolor=\color{background},
    literate=
     *{0}{{{\color{numb}0}}}{1}
      {1}{{{\color{numb}1}}}{1}
      {2}{{{\color{numb}2}}}{1}
      {3}{{{\color{numb}3}}}{1}
      {4}{{{\color{numb}4}}}{1}
      {5}{{{\color{numb}5}}}{1}
      {6}{{{\color{numb}6}}}{1}
      {7}{{{\color{numb}7}}}{1}
      {8}{{{\color{numb}8}}}{1}
      {9}{{{\color{numb}9}}}{1}
      {:}{{{\color{punct}{:}}}}{1}
      {,}{{{\color{punct}{,}}}}{1}
      {\{}{{{\color{delim}{\{}}}}{1}
      {\}}{{{\color{delim}{\}}}}}{1}
      {[}{{{\color{delim}{[}}}}{1}
      {]}{{{\color{delim}{]}}}}{1},
}

\usepackage[framemethod=tikz]{mdframed}

\usepackage{xspace}
\usepackage{comment}
\usepackage{url}
\usepackage{subcaption}

\usepackage[utf8]{inputenc}

\usepackage{algorithm}
\usepackage[commentColor=blue]{algpseudocodex}

\title{Performance and competence intertwined: A computational model of the Null Subject stage in English-speaking children}
\author{Soumik Dey \\
  The Graduate Center \\
  The City University of New York \\
  \texttt{sdey@gradcenter.cuny.edu}\\\And
  William Gregory Sakas \\
  The Graduate Center \\
  The City University of New York \\
  \texttt{wsakas@hunter.cuny.edu}}

\date{}

\begin{document}

\maketitle



\begin{abstract}
The empirically established null subject (NS) stage, lasting until about 4 years of age, involves frequent omission of subjects by children. \citet{10.2307/23358022} observe that young English speakers often confuse imperative NS utterances with declarative ones due to performance influences, promoting a temporary null subject grammar. We propose a new computational parameter to measure this misinterpretation and incorporate it into a simulated model of obligatory subject grammar learning. Using a modified version of the Variational Learner \cite{yang2012} which works for superset-subset languages, our simulations support Orfitelli and Hyams' hypothesis. More generally, this study outlines a framework for integrating computational models in the study of grammatical acquisition alongside other key developmental factors.

\end{abstract}

\section{Introduction}

The Null Subject (\textit{NS}) stage is a well-researched phenomenon in child language acquisition, characterized by young children sometimes forming declarative sentences without subjects. This is expected in children exposed to null subject languages but contentious in obligatory subject language environments. The NS stage challenges the Subset Principle \cite{GOLD1967447,10.7551/mitpress/1074.001.0001,10.2307/4178549, VALIAN1990105, 10.2307/4178801, fodor_sakas_2005} ---  children learning obligatory subject languages exhibit NS-like sentences (a superset language), which gradually shift to non-NS (subset language) with time. This phenomenon puzzles learning theorists. Explanations vary, with some attributing the NS stage to differences between children's internal grammar and adult target grammars \cite{yang2012, orfitelli2008experimental, VALIAN1990105}, while others cite extrasyntactic factors like memory and processing constraints \cite{Bloom1970,10.2307/4178692,VALIAN199121,wang1992null}. \citet{rizzi2005a,rizzi2005b} connects a performance account of a limited production system with its consequence of the varying grammatical competence we see in children. In this paper, we model the grammatical theory of the NS stage in children using a developmental parameter and the Variational Learner (VL)\cite{yang2012}, a well-known computational model of language acquisition. More generally, this study outlines a framework for integrating computational models in the study of language acquisition alongside other key developmental factors.

\section{Background}
\subsection{Orfitelli \& Hyams (2012) Experiment 2 \label{secOH}}

The two distinct theories of performance and grammatical competence present distinct explanations for children's comprehension of subject-lacking sentences (\textit{NS sentences}, such as imperatives in English). Grammatical theories propose that young English speakers view NS sentences akin to grammatically correct declaratives, similar to adults in null subject languages. Conversely, performance theories attribute omissions to production limitations, suggesting children interpret NS sentences as adults do in obligatory subject languages, limiting English-speaking children's interpretations to imperatives or diary forms. To explore this, \citet[Experiment 2]{10.2307/23358022} used a truth-value judgment (TVJ) experiment \cite{crain1985acquisition,crain1993}. In Experiment 2, a child watched a narrative, then listened to a puppet's (Mr. Bear) comment, and judged the comments' accuracy relative to the story. The child corrected Mr. Bear by indicating the correctness of his statements, with explanations. Thirty children from Los Angeles daycare centers were involved, divided into three age groups (2;6-2;11, 3;0-3;5, and 3;6-3;11) to represent early, middle, and late NS stages. Details on age range and distribution are in Table \ref{ohtable}, adapted from \citet[Table 6]{10.2307/23358022}.

\begin{table}[t]
\centering
\begin{tabular}{@{}cccc@{}}
\toprule
Group    & Age Range & Mean Age & N  \\ \midrule
2;6-2;11 & 2.54-2.96 & 2.73     & 10 \\
3;0-3;5  & 3.12-3.48 & 3.3      & 10 \\
3;6-3;11 & 3.64-3.98 & 3.82     & 10 \\ \midrule
Total    & 2.54-3.98 & 3.28     & 30 \\ \bottomrule
\end{tabular}
\caption{\citet{10.2307/23358022}[Experiment 2] participant details.} 
\label{ohtable}
\end{table}

The children underwent assessment on 24 grammatical items (sentences), equally split between correct and incorrect true/false responses. There were 8 NS condition sentences, while the remaining 16 items were evenly divided among the remaining four conditions. Orfitelli \& Hyams (O\&H) classified the children's responses to NS condition sentences into three categories based on interpretation:
\begin{itemize}
    \item \textbf{Consistently imperative}: 7-8 out of 8 NS sentences interpreted as imperative.
    \item \textbf{Both interpretations allowed}: 2-6 out of 8 NS sentences interpreted as imperative.
    \item \textbf{Consistently declarative}: 0-1 out of 8 NS sentences interpreted as imperative.
\end{itemize}

\begin{table}[h]
\centering
\begin{tabular}{@{}c||ccc@{}}
\toprule
                                                                 & \textbf{2;6-2;11} & \textbf{3;0-3;5} & \textbf{3;6-3;11} \\ \midrule
\begin{tabular}[c]{@{}c@{}}Imperative\\ (7-8 imp)\end{tabular}   & 0\% (0)           & 40\% (4)         & 80\% (8)          \\ \midrule
\begin{tabular}[c]{@{}c@{}}Both \\ (2-6 imp.)\end{tabular}       & 80\% (8)          & 60\% (6)         & 20\% (2)          \\ \midrule
\begin{tabular}[c]{@{}c@{}}Declarative\\ (0-1 imp.)\end{tabular} & 20\% (2)          & 0\% (0)          & 0\% (0)           \\ \bottomrule
\end{tabular}
 \caption{Individual performance on the NS condition sentences in \citet{10.2307/23358022}[Table 8].}
 \label{ohtable2}
\end{table}

\begin{figure}[t]
\centering
\includegraphics[width=0.5\textwidth]{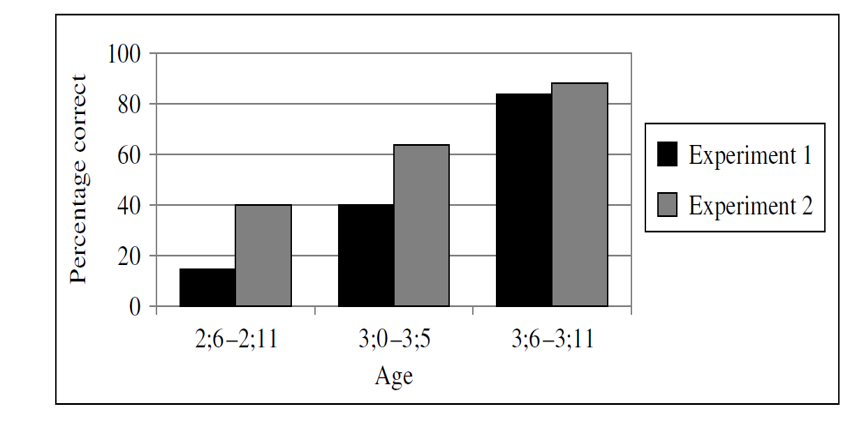}
\caption{Performance on NS condition sentences from \citet{10.2307/23358022}[Figure 5].}
\label{ohfig}
\end{figure}

Additional details regarding the performance of children on the task are illustrated in Figure \ref{ohfig}. The performance on the NS items varied with age as O\&H reported that the youngest group assigned an imperative (adult) interpretation to NS items 40\% of the time on average, while the middle age group assigned an imperative interpretation 64\% of the time. While O\&H do not report an average number for the performance on NS items for the oldest age group, from Figure \ref{ohfig} we can estimate the average performance to be close to 90\%. For concreteness, we adopt 90\% for this age group from this point on.

 The fact that in O\&H's study children comprehend NS sentences differently than an adult, reinforces the grammatical account of the NS stage. However, O\&H also argue for performance limitations which create the illocutionary force ambiguity associated with the imperative NS sentences resulting in the NS stage.

\subsection{Language Acquisition in P\&P Framework}

The study of language acquisition presents an extraordinary challenge for scientific inquiry. It requires that a child, over a remarkably short period, must develop a grasp of a grammar system capable of producing and interpreting a set of utterances comparable to those produced by adults within their linguistic surroundings.\footnote{Or nearly identical, encompassing microvariations within linguistic communities.} The child's cognitive mechanisms for language learning achieve this despite having limited or no exposure to sentence-level linguistic phenomena, and without the capacity to perceive intrinsic properties of the latent structures that generate the surface forms of utterances (see \citealt{fodor_unambiguous_1998} for reference). This restricted interaction with sufficient surface forms forms the core of the argument known as the poverty of the stimulus \cite{chomsky_lectures_1981, chomsky_logical_1955, chomsky_aspects_1965, chomsky_knowledge_1986}. 

This argument has been instrumental in advocating for an intrinsic language faculty that imparts universal structural principles (such as the notion that all languages possess subjects) and parameters, which dictate language-specific structural traits (e.g., whether SpecIP is initial or final) that are adjusted during the language acquisition process.,The framework of principles and parameters (P\&P), as introduced by \cite{chomsky_lectures_1981}, was designed to streamline the language learning process by:

\begin{itemize}

\item Limiting the potential scope of grammatical possibilities, transitioning linguistic theory from a potentially limitless universe of human grammars to a explicitly finite set
\item Simplifying complex structural phenomena into parameter values, which vary between languages. This framework comprises a set of foundational principles that “sharply restrict the class of
attainable grammars and narrowly constrain their form, but with parameter [values] that have to be fixed by experience” \cite{chomsky_lectures_1981}. 

\end{itemize}

Essentially, the child is inherently equipped with these principles, while parameter values are influenced by the linguistic input they encounter in their environment.\footnote{For the sake of simplicity, linguistic learnability typically presumes the language environment as monolingual.} We align with Fodor's interpretation of parameter values within the P\&P model (\citealt{fodor_sakas_2005}): Universal Grammar (UG) endows parameters with two possible, albeit mutually exclusive, structural "treelets" -- elements of grammatical architecture -- that serve as tools for both linguists and children acquiring language to distinguish between different human languages. Subsequently, \cite{doi:10.1080/10489223.2020.1803329} suggest that parameter values should be seen \textit{not} as simple binary choices between parametric treelets, but rather as points within a gradient spectrum between these discrete choices, viewing parameter values as dynamically adjustable along a continuum.

\subsection{Variational Learner\label{secVL}}

\citet[34]{RN73} argues for the necessity of learners to perform well in domains without unambiguous inputs (see \citealt{RN19,RN75} who argue against the general existence of unambiguous evidence). He proposes a parameter setting reward-based algorithm that converges to a target grammar despite the presence of ambiguous evidence \cite{strauss2008}. His \textit{Variational Learning} (\textit{VL}) model posits that a child accesses multiple grammars, competing throughout learning. When encountering a sentence, the child uses her current grammar hypothesis for parsing. Success results in rewards; failure incurs penalties. Competing grammars vie to become the next hypothesis, with the most rewarded becoming the adult grammar. In VL, a \textit{learning rate} $\hat{R}$ dictates grammar rewards or penalties. Each grammar $G_i$ is linked to a probability $P_i$, indicating past rewards and penalties. At time $t$, this probability $P_i$ depends on linguistic exposure $E_t$ and grammar performance. Implementing variational learning with Principles and Parameters involves managing \textit{$2^n$} probabilities for grammars in an \textit{n}-parameter space, exceeding a billion in a 30-parameter P\&P domain. Yang suggests maintaining one weight ($w_i$) per parameter ($p_i$). Like non-parametric VL, parameters are adjusted based on parsing outcomes, modifying weight ($w_i$) accordingly. Each $p_i$ is binary, with value ($p_i^v$) of 0 or 1. Grammar probabilities form a weight vector ($W$) of size $n$, where $w_i$ aligns with parameter $p_i$. Weights encode cumulative parametric reward and penalty results at time $t$ after $E_t$. In P\&P VL, \citet{RN73} details two weight update methods following a sentence parse ($s_t$) at time $t$. Weights vector $W=[w_1, w_2 ... w_n]$ is adjusted, rewarding successful parsing by $G_{curr}=[p_1^v, p_2^v ... p_n^v]$ and penalizing failures. Updated weights then define new $G_{curr}$. \citet{yang2012} describes a \textit{reward-only} VL where unsuccessful parsing leaves weights unaltered. Following \citet{sakas2017parameter}, we adopt and modify principles and parameters reward-only VL for simulations.

The reward scheme of the reward-only VL follows the ($L_{R - P}$) scheme of \citet{RN259}. If a parameter value, $p_i^v$, in $G_{curr}$ is 0 and $w_i$ is to be rewarded, the weight is nudged towards 0 according to Equation (\ref{VLreward0}):
	 \begin{equation}\label{VLreward0}  w_{i}^{t+1} = w_{i}^{t} - \hat{R} \cdot(w_{i}^{t}) \end{equation}
	 
If a parameter value, $p_i^v$, in $G_{curr}$ is 1 and $w_i$ is to be rewarded, the weight is nudged towards 1 according to Equation (\ref{VLreward1}):
\begin{equation} \label{VLreward1} w_{i}^{t+1} = w_{i}^{t} + \hat{R}\cdot(1-w_{i}^{t}) \end{equation}

Where $w_{i}^{t}$ denotes weight $w_i$ in the vector of weights $W$ at time instance $t$. $w_{i}^{t+1}$ is the weight after the update when encountering the input sentence $s_t$ at time instance $t$.

\citet{yang_2002} hypothesized that a child is unable to distinguish between English grammar and its NS counterpart  early on (imperfect learning), while in later stages of acquisition the corrective force of grammar competition sets the target parameter correctly.






\section{Computational Modelling of NS utterance interpretation}

The remainder of this paper presents a simulation study which models the work of \citet{10.2307/23358022}. Specifically, we model the increase of a child's ability to interpret imperative sentences in an adult manner and observe the change in a simulated learner's (an \textit{e-child}'s) competence over the course of language acquisition in an English-like abstract linguistic environment. We run simulation experiments employing a computational models of syntactic parameter setting: The Variational Learner \cite{RN73,yang2012,sakas2017parameter} which we modify to incorporate (a version of) the Subset Principle. These experiments are run on the English-like language drawn from a large domain developed at the City University of New York (CUNY). The study presents a computational investigation of how performance factors might influence competence longitudinally.

\subsection{The CUNY-CoLAG language domain \label{seccolag}}

The CUNY-CoLAG domain is a database of word order patterns that children could be expected to encounter, together with all syntactic derivations of those patterns and the syntactic parameter values which generated each derivation. The multi-language domain is large, containing 3,072 artificial languages, 48,077 distinct word order patterns, and 93,768 distinct syntactic trees. Germane to this article, is CoLAG English \cite{sakas2017parameter}, most English-like language in the domain. A more thorough overview of the domain and how the multilingual derivations were generated can be found in \citet{sakas-2003-word} and the most recent version of the CUNY CoLAG domain (hereafter, simply CoLAG) is comprehensively presented in \citet{sakas2011generating,sakas2012disambiguating}.\footnote{The domain is available for download at: \url{https://bit.ly/3nGdhPc}.} \footnote{While we acknowledge that CoLAG is an artificial domain, natural language domains like CHILDES\cite{RN239} has been explored with the VL in \cite{sakas2017parameter}. The CoLAG domain is used to prove a theoretical point and test the convergence pattern of model with a wide variety of distributions reminiscent to the study in \cite{doi:10.1080/10489223.2020.1803329}.}

\subsection{Subset-superset languages and the Variational Learner\label{secSSVL}}

Yang's Variational Learner is highly regarded in terms of bringing statistical methods to the table together with generative grammar. However, the VL cannot distinguish between superset and subset grammars and cannot be prevented from converging on an incorrect superset hypothesis. However, a version of Yang's learner that \textit{does} distinguish between superset and subset grammars and avoid convergence on an incorrect superset hypothesis can be envisioned: Whenever the learner encounters a sentence licensed by a current grammar hypothesis which generates a superset language, it checks if the sentence can be parsed by a subset hypothesis of the current grammar. If the sentence can be parsed by the subset grammar the learner picks the subset grammar choice, rather than the current (superset) grammar hypothesis for adjusting the weights (Yang, p.c.).

We embrace this strategy, however, we found a need to augment it. The strategy focuses on acquiring a target subset grammar and is potentially detrimental, in the worst case fatally, when the VL is faced with a superset target grammar. Suppose an e-child employing the VL is trying to learn a superset target grammar. Every time the e-child hears an utterance that can be parsed by the subset grammar, the learner adjusts its weights in the direction of the subset grammar. Thus, convergence towards the superset is dependent on the order, and the ratio of sentences unambiguously licensed by the superset grammar to those licensed by the subset grammar.

 \begin{figure}[t]
\centering
\includegraphics[width=0.5\textwidth]{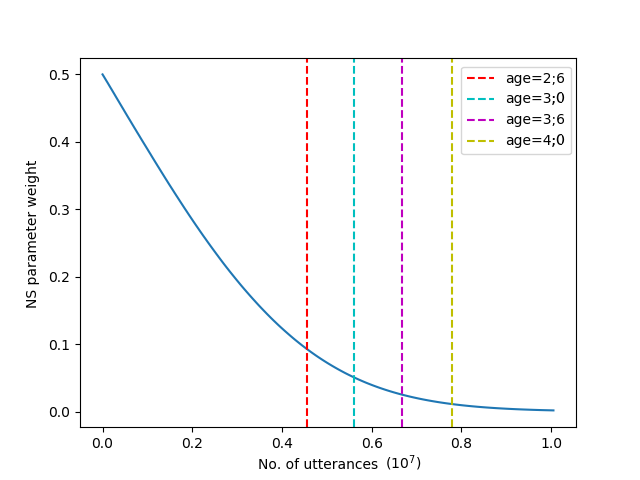}
\caption{The SSVL with a conservative learning rate of $r=1.24\times 10^{-7}$. The NS parameter weight is plotted on the y-axis and the number of utterances on the x-axis. Additionally, 6 month intervals from age 2;6 to 4;0 as measured in number of utterances are marked.}
\label{vanillaVL}
\end{figure}

To confirm our suspicions we ran this version of the Variational Learner with a 100 e-children acquiring CoLAG Null Subject English (NS-English), i.e., a language that has all the CoLAG English parameter settings except for Null Subject which has a value of 1 allowing null subjects in declaratives. All 100 e-children converged incorrectly on the subset value of the Null Subject parameter.\footnote{\label{fcons}Following \cite{sakas2017parameter}, the simulations were run on a uniform distribution of CoLAG English sentences with a learning rate of $0.001$ and successful convergence was defined as the weights reaching within $0.02$ threshold of the target parameter values.} 

We propose an adaptation of the VL which allows it to consistently converge to the correct parameter setting of a superset-subset parameter. The approach we adopt is to ensure that whenever the VL encounters a sentence that can be parsed only by the superset grammar, we reward it at a higher rate in comparison to the rate used for the subset value. The idea is to have two learning rates --- a higher rate for rewarding the superset and a lower rate for rewarding the subset. Following \citet{doi:10.1080/10489223.2020.1803329}, we will call them the ``aggressive" ($R$) and ``conservative" learning rates ($r$) respectively. To test this idea, we again ran simulations involving a 100 e-children acquiring CoLAG NS-English with learning rates of $R=0.008$ and $r=0.001$.\footnote{In line with Footnote \ref{fcons}, the aggressive rate of $0.008$ was chosen through trial and error for the learner to converge with a conservative rate of $0.001$.} This learning strategy is successful \textemdash{} the \textit{Superset-Subset Variational Learner} (\textit{SSVL}) successfully converges on the target superset value for the Null Subject parameter for all e-children acquiring CoLAG NS-English, see Figure \ref{vanillaVL}.

\begin{algorithm}[t]
\small
    $W$ is the array of weights\;
    $G_{curr}$ is the current grammar, i.e. vector of parameter values\;
    \\
    $n$ is the number of parameters\;\\
	\SetAlgoLined
	\For{each $w_i$ in W}{
		$w_i \gets 0.5$ \;
	}
	\For {each input sentence s}{
	    pick $G_{curr} \gets$ [$p_1^v$, ... ,$p_n^v$] according to Algorithm \ref{VL}\;
	   	\uIf {$G_{curr}$ can parse $s$}{
		    \For {$w_i$ in $W$}{
		          \uIf{$p_i$ is a not superset-subset parameter or $p_i^v$ is the subset value}    {
		                Adjust $w_i$ conservatively towards $p_i^v$\; 
		          } 
		          \Else(\tcp*[h]{$p_i^v$ is the superset value})
		               {$G_{temp} \gets $[$p_1^v$, ...,$1-p_i^v$, ... $p_n^v$]\hspace{.3cm} \tcp*[h]{$1-p_i^v$=subset value}
		          		
		                    		                    \uIf{ $G_{temp}$ can parse $s$}{ Adjust $w_i$ conservatively towards $1-p_i^v$;
		                  }
		                  \Else{
		                  Adjust $w_i$ aggressively towards $p_i^v$;
		                  }
                         }
                    }
		        }
		  }
\caption{Superset-Subset Yang's Variational Learner reward only.}
\label{SSVL}
\end{algorithm}

Pseudocode for the SSVL is given in Algorithm \ref{SSVL}. The initialization of the weights and the pick of $G_{curr}$ is identical to Algorithm \ref{VL} (Yang's Reward-only VL). After every sentence,  if the input sentence can be parsed by $G_{curr}$, the SSVL checks all $p_i^v$ in $G_{curr}$ for superset-subset values, if any. If $p_i$ is not a superset-subset parameter 
or if $p_i$ is a superset-subset parameter and $p_i^v$ is the subset parameter value, $w_i$ is rewarded conservatively towards $p_i^v$. Otherwise, $p_i^v$ is a superset value. In that case, the SSVL checks if the current grammar with the superset value of $p_i$ flipped ($G_{temp}$ in Algorithm \ref{SSVL}) to the subset value $1-p_i^v$ can parse the current input sentence, if so, $w_i$ is rewarded conservatively towards the subset value, Otherwise $w_i$ is rewarded aggressively towards the superset value. As with the original reward-only VL, if the current input sentence can not be parsed by $G_{curr}$ no weight updates occur. 

The weights in $W$ are rewarded as follows: 

\begin{itemize}
    \item \textbf{Reward aggressively}: Replace $\hat{R}$ by the aggressive rate $R$ in Equation (\ref{VLreward0}) or (\ref{VLreward1}) and update $w_i$ accordingly.
    \item \textbf{Reward conservatively}: Replace $\hat{R}$ by the conservative rate $r$ in Equation (\ref{VLreward0}) or (\ref{VLreward1}) and update $w_i$ accordingly. 
\end{itemize}

The original Variational Learner follows the Naive Parameter Learning (NPL) model, which assumes that when the composite grammar successfully parses the incoming sentence, all parameter values are rewarded. However, as seen in our experiments involving CoLAG English and NS-English, for successful convergence on either the superset or subset grammar, the VL cannot not afford to be ``naive''. Specifically, it requires knowledge of which parameter values are in superset-subset relationship and exactly how to reward the relevant value.

\subsection{A performance parameter: IARC}

\citet{10.2307/23358022} based on their TVJ experiment, observe that there is a misinterpretation of illocutionary force in null subject sentences due to performance limitations in children and conjecture that adults and children have different grammars. This section outlines the modeling approach we adopt to capture this observation.

Table \ref{ohtable2} presents data that show that children's ability to correctly interpret imperative illocutionary force changes over time. This change is almost linear: Children between ages 2;6-2;11 show 40\% adult interpretation of NS utterances on average, while children between ages 3;0-3;5 show 64\% and 3;6-3;11 show 90\%.

`We computationally model the interpretation of the illocutionary force of an e-child by introducing imperative NS sentences labeled with declarative illocutionary force into the e-child's linguistic environment. This ``\textit{noisy}'' input to an e-child can be manipulated to mirror the data in Table \ref{ohtable2} by decreasing the noise as the e-child matures. Employing this simulated performance factor, we map the pathway the NS parameter takes during the acquisition of CoLAG English. The question we are asking is --- Assuming English children do indeed have a declarative interpretation of imperative NS sentences \textemdash{} how can we model the change in the Null Subject parameter, a parameter whose acquisition is affected by these NS sentences, to come to a conclusion regarding its target setting? And given the projected trajectory of this developmental change, what course would the trajectory of NS parameter acquisition take?

In learning CoLAG English, reliance must be placed on declarative utterances with subjects. Misunderstanding imperative NS forms as declaratives impairs learning, treating some NS utterances as noise and incorrectly shifting the parameter toward the null subject superset. In obligatory subject languages for adults, mature children's transition should reflect a shift from superset (optional subject) to subset (obligatory subject) grammars. Drawing on \citet{10.2307/23358022}'s TVJ experiment, we introduce the \textit{Illocution Ambiguity Resolution Coefficient} (IARC) for measuring children's misinterpretations of imperatives. An IARC of 1 indicates perfect recognition of imperative NS as such, whereas an IARC of 0 signifies total misinterpretation as declaratives. An IARC of 0.2 suggests 20 out of 100 NS imperatives are correctly understood, with 80 misunderstood as declaratives. Our goal is to explore NS parameter acquisition in CoLAG English, considering these performance limitations.

\subsection{Growth of IARC}

In this section, we develop a framework for quantifying how the performance parameter IARC grows as a function of age, measured here by the cumulative utterances heard by a child at the end of age range $i$ ($U_i$). Recall that $IARC$ is a probability measure and hence is bound within the values 0 and 1. As discussed in Section \ref{secOH},  the average values of IARC between the age ranges of 2;6-2;11, 3;0-3;5 and 3;6-3;11 exhibit almost linear growth (0.4 to 0.64 to 0.9). Thus, a natural way to model IARC would be as a bound function of $U_i$, $0\leq IARC\leq 1$ with IARC linearly increasing with respect to $U_i$. One such approach is presented in Equation (\ref{linear}), where $m$ is the slope, and $c$ is the intercept of a linear function of IARC growth.

\begin{equation}
    IARC_{\text{linear}}(U_i) =
    \begin{cases}
        0 & U_i \leq -\frac{c}{m} \\
        m U_i + c & -\frac{c}{m} \leq U_i \leq \frac{1 - c}{m} \\
        1 & U_i \geq \frac{1 - c}{m}
    \end{cases}
    \label{linear}
    \vspace{4mm}
\end{equation}

In addition to the $IARC_{linear}$ function, we also employ a logistic function implementation of IARC, $IARC_{logistic}$ as shown in Equation  (\ref{logistic}) bound by 0 and 1, with growth rate $m$ and midpoint $c$. The logistic function exhibits an s-shaped (sigmoid) curve. For a sufficiently low $m$, the logistic function behaves almost linearly across the midpoint $c$ and is asymptotic at the (0 and 1) endpoint values. 

\begin{equation}
    IARC_{logistic}(U_i) = \frac{1}{1+e^{-m\times(U_i-c)}}
    \label{logistic}
    \vspace{4mm}
\end{equation}

\subsection{Simulation of a 100 e-children\label{sec100child}}

\begin{figure*}[t]
    \begin{subfigure}{0.5\textwidth}
    \includegraphics[width=\textwidth]{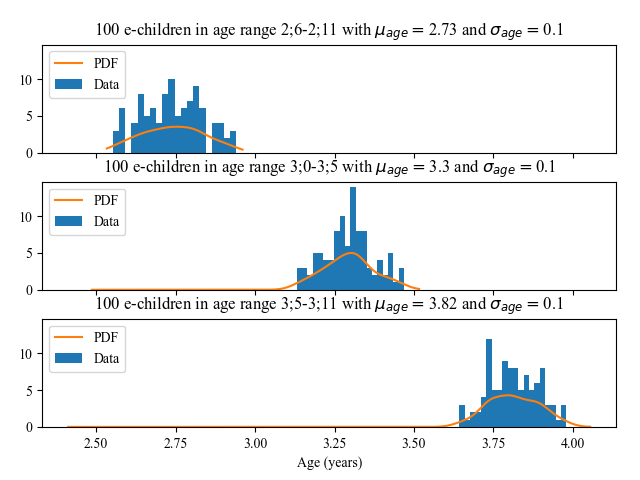}
    \caption{Frequency histogram of ages.}
    \label{age_dist}
    \end{subfigure}
    \hfill
    \begin{subfigure}{0.5\textwidth}
    \includegraphics[width=\textwidth]{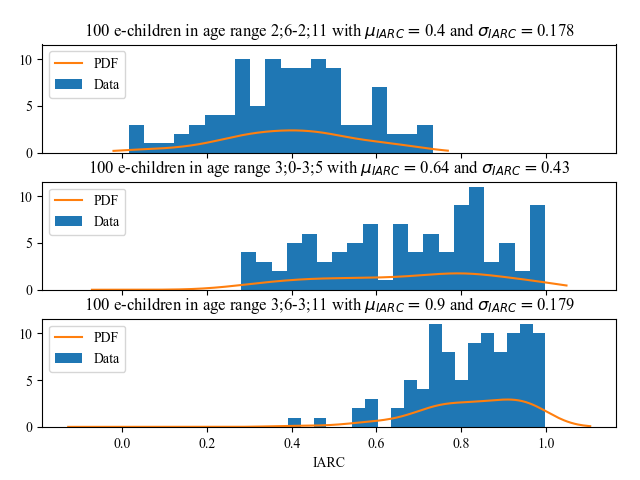}
    \caption{Frequency histogram of IARC values.} 
    \label{IARC_dist}
    \end{subfigure}
    \caption{Performance of 100 e-children with a bin size of 20 and the Gaussian kernel estimation of the probability density function (\textit{PDF}) across 3 age groups generated using a truncated Gaussian distribution emulating O\&H.}
\end{figure*}

Building on the research presented in \citet{doi:10.1080/10489223.2021.1888295, hart1995meaningful, hart2003early}, our estimate is that by age 5;0, a child from a professional-class background has been exposed to 10,054,267 utterances. To depict the variability among children noted in O\&H's Experiment 2, we simulate 100 virtual children using a truncated Gaussian age distribution for each age category listed in Table \ref{ohtable}, which specifies the minimum, maximum, and mean ages. Constructing a truncated Gaussian demands parameters such as range, mean ($\mu$), and standard deviation ($\sigma$). O\&H provide age ranges and mean ages ($\mu_{age}$) in Table \ref{ohtable}, but omit standard deviations for each group ($\sigma_{age}$). We approximate these standard deviations ($\sigma_{age}$) as 0.1 for all age groups based on available age ranges. According to \citet{gleitman1984current}, imperatives make up about 16\% of the language input a child receives until age 2. Earlier, \citet{ngn1977} estimated an 18\% imperative usage beyond age 2. With $IARC=0$, this reflects the expected level of noise (imperatives misunderstood as declaratives) encountered by the learner. Our simulations modulate the IARC parameter following Equations (\ref{linear}) or (\ref{logistic}), assuming an imperative exposure rate of 16\% up to age 2 (approximately 3,566,210 utterances) and 18\% thereafter.

\begin{table}[t]
\small
\centering
\begin{tabular}{@{}c||ccc@{}}

\toprule
                                      & \textbf{2;6-2;11}                                                                   & \textbf{3;0-3;5}                                                                    & \textbf{3;6-3;11}                                                                   \\ \midrule
IARC Range                            & 0-0.75                                                                     & 0.25-1                                                                     & 0.25-1                                                                     \\ \midrule
\multirow{2}{*}{Tail end probability} & \multirow{2}{*}{\begin{tabular}[c]{@{}c@{}}($<0.25$)\\ = 0.2\end{tabular}} & \multirow{2}{*}{\begin{tabular}[c]{@{}c@{}}$(>0.75$)\\ = 0.4\end{tabular}} & \multirow{2}{*}{\begin{tabular}[c]{@{}c@{}}$(<0.75$)\\ = 0.2\end{tabular}} \\
                                      &                                                                            &                                                                            &                                                                            \\ \midrule
$\mu_{IARC}$                          & 0.4                                                                        & 0.64                                                                       & 0.9                                                                        \\ \midrule
$\sigma_{IARC}$                       & 0.1785                                                                     & 0.43                                                                       & 0.179                                                                      \\ \bottomrule
\end{tabular}
\caption{Table depicting the calculation of standard deviation of the Gaussian distribution of IARC parameter over age ranges.}
\label{tail_end}
\end{table}

Similar to the age data, O\&H do not provide the standard deviation of IARC ($\sigma_{IARC}$) for the 3 age groups. However, O\&H do provide some additional distributional data which can be used to estimate the standard deviation. The O\&H IARC data, presented in Table \ref{ohtable2}, has been recast as distributional metrics in Table \ref{tail_end}. We infer the tail end probabilities of the IARC distribution from Table \ref{ohtable2} in order to calculate the standard deviation ($\sigma_{IARC}$) of each age group. We observe that for ages 2;6-2;11, 20\% of the children correctly interpret less than 2 out of 8 imperatives (IARC value less than 0.25), i.e., the probability that IARC is less than 0.25, $P(IARC)<0.25$, is 0.2 (20\% of the children). In addition, the mean IARC ($\mu_{IARC}$) of this age range is 0.4 as reported in Section \ref{secOH}. With this tail end probability and the mean ($\mu_{IARC}$), the standard deviation of the Gaussian distribution of the children's IARC value ($\sigma_{IARC}$) for the age range 2;6-2;11 was estimated to be 0.1785. The tail end probabilities for the other age groups were similarly inferred\footnote{For the middle age group, the right tail was used rather than the left.} and the standard deviations of all three age ranges are calculated and compiled in Table \ref{tail_end}.

Using the parameters discussed above, a truncated Gaussian distribution in Scipy was used to generate an age distribution and an IARC distribution using two growth functions \textemdash{} $IARC_{linear}$ and $IARC_{logistic}$, of a 100 e-children across 3 age groups as depicted in Figures \ref{age_dist} and \ref{IARC_dist} respectively. After a 100 age and IARC values for each of the three age groups were generated, we sort the ages and the IARC values within each group. To approximate the longitudinal development of the IARC value for each e-child in the pool of a 100 e-children, we generate three $(IARC, age)$ pairs for each e-child, one from each age group, using the sorted IARC and age values. The first $(IARC, age)$ value in each of the three lists is used to generate e-child 1, the second three $(IARC, age)$ pairs are used to generate e-child 2, etc. Using these three $(IARC, age)$ pairs for each e-child, the optimal parameters for the two growth functions \textemdash{} $IARC_{linear}$ and $IARC_{logistic}$ \textemdash{} were calculated as outlined previously in this section. We then proceed to simulate the acquisition of the NS parameter for each of the resulting 100 e-children.

 \begin{algorithm}[t]
 \small
\SetAlgoLined
            $G_{targ}$ is the target grammar\;
            $IARC$ is the probability of interpreting an imperative sentence correctly as an imperative \;\\
            $m$ and $c$ are the optimal parameters for $IARC$ growth\;\\
            $G_{targ} \gets$ CoLAG English, i.e., 0001001100011\\
            $IARC \gets 0$\;\\
            $num\_sentences \gets$ 10,054,267, i.e., cumulative utterances by age 5;0\;\\
            \For {i in range($num\_sentences$)}{
            calculate IARC using $m$ and $c$ accordingly\;
            \If{$i<3,566,210$, i.e., $age<2;0$}{$s \gets$ sentence from the target language with 16 percent probability of being an imperative\;}
            \Else{$s \gets$ sentence from the target language with 18 percent probability of being an imperative\;}
            \If{$s$ is imperative}{
            with probability of, 1-IARC, interpret $s$ as a declarative\;
            }
            Run SSVL on $s$\;
            }
 \caption{Simulation of one e-child incorporating IARC.}
 \label{TVJ1}
\end{algorithm}

We conducted 2 experiments with a pool of 100 e-children employing the SSVL acquiring CoLAG English with 2 growth functions $IARC_{linear}$ and $IARC_{logistic}$. The e-children were generated according to the methodology described above. The simulations used an aggressive rate ($R$) of $2\times 10^{-4}$ and a conservative rate ($r$) of $5\times 10^{-6}$. \footnote{\label{chlearn}The choice of learning rates was derived through trial and error.} The NS parameter weight / confidence values of these 100 e-children over time are plotted on the y-axis of the graphs presented in Figures \ref{VLlinear100} and \ref{VLlogistic100} with the x-axis representing the number of cumulative utterances encountered. To show the variation of the e-children, all 100 are plotted with the fastest, the slowest, and the median e-child, in terms of convergence speed, demarcated.

\section{Results\label{secres}}
\begin{figure*}[t]
    \centering
    \begin{minipage}{0.5\textwidth}
        \centering
        \includegraphics[width=\linewidth]{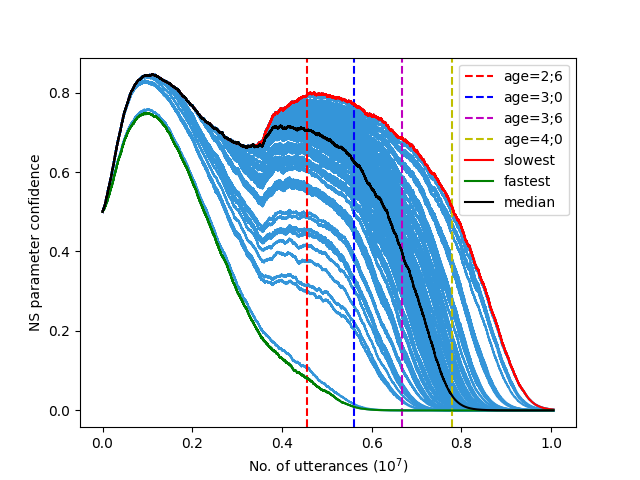}
        \subcaption{$IARC_{linear}$ growth function}
        \label{VLlinear100}
    \end{minipage}\hfill
    \begin{minipage}{0.5\textwidth}
        \centering
        \includegraphics[width=\linewidth]{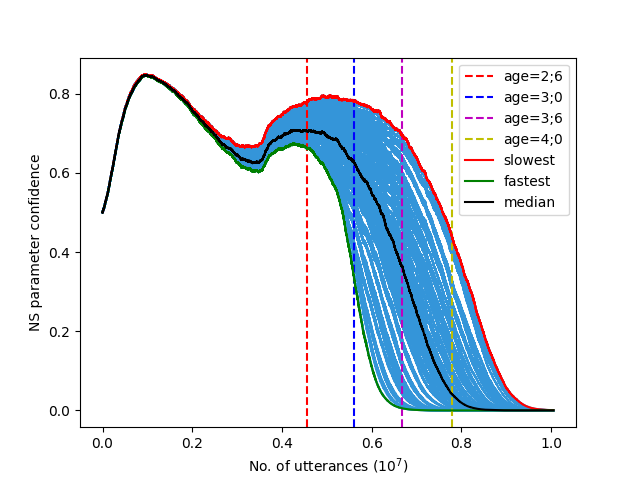}
        \subcaption{$IARC_{logistic}$ growth function}
        \label{VLlogistic100}
    \end{minipage}
    \caption{The SSVL employing the  with learning rates $R=2\times 10^{-4}$, and $r=5\times 10^{-6}$ for 100 e-children. The fastest, the slowest and the median e-children, in terms of convergence speed, are highlighted. } 
\end{figure*}

In the work of \citet{10.2307/23358022}, a significant empirical discovery regarding developmental constraints is presented, specifically focusing on the differential interpretation of Null Subject (NS) sentences between adults and children. The NS stage arises from an intricate interplay of grammatical and performance elements. The objective of the study's simulations is to replicate this early-stage developmental constraint and the ensuing partial learning observed over time. The objective was to create electronic children, or e-children, whose linguistic development could accurately reflect the longitudinal findings of O\&H. The simulations were conducted using a specifically adapted variational learning model (SSVL) incorporating superset and subset language frameworks, alongside two distinct models of IARC growth ($IARC_{linear}$ and $IARC_{logistic}$). The experiments with the SSVL model (illustrated in Figures \ref{VLlinear100} and \ref{VLlogistic100}) reveal a particular behavior of the Null Subject parameter: it begins at an initial value of 0.5, then swiftly ascends to approximately 0.8, before subsequently declining, which mirrors the observed decrease in the employment of null subjects among English-speaking children. A minor resurgence occurs around age 2;0 due to a simulated increase in imperative sentence exposure experienced by an e-child, as elaborated in Section \ref{sec100child}. This phenomenon is supported by findings from two distinct studies on imperatives directed at young children, one before the age of 2;0 and the other thereafter. Furthermore, we also performed simulations of the SSVL model in a noiseless setting within the CoLAG environment acquiring NS-English. Under noiseless conditions, SSVL demonstrates that the NS parameter promptly converges well before the e-children reach 2;0, the age traditionally associated with the onset of the NS stage.
\section{Summary and Discussion}

\citet{10.2307/23358022} observe that young English-speaking children often misinterpret (subjectless) imperative utterances as declaratives (e.g., \textit{Play with blocks.}), which could potentially lead them to initially acquire an NS grammar. The present study computationally models the findings of \citet{10.2307/23358022}.
More generally, it establishes a framework for simulating a developmental \textit{‘performance parameter’} and its influence on acquisition. The performance parameter relevant to \citet{10.2307/23358022} and the computational work reported here we coin the \textit{Illocution Ambiguity Resolution Coefficient} (IARC) — a measure of a child’s ability to correctly disambiguate between imperative and declarative illocutionary force in utterances without a subject.

We computationally model the performance parameter IARC, based on empirical data from \citet{10.2307/23358022}, and study its effect during acquisition of the NS syntactic parameter. Employing a modified version of the  \textit{Variational Learner} (VL, \citealt{RN73,yang2012}), we simulate the change over time in the confidence value associated with the NS parameter in simulated ‘e-children’ acquiring an English-like language in an artificial language domain \cite{sakas2011generating}.
The VL cannot reliably learn languages in superset/subset relationships \citet{sakas2017parameter}, which is critical to modeling the acquisition of the NS parameter. To employ the VL paradigm in this context, we develop the \textit{Superset/Subset Variational Learner} (SSVL) \textemdash{} a version of the VL that can effectively distinguish superset and subset grammars and successfully acquire them.

Simulating 100 SSVL e-children employing two growth functions of IARC, we observe that the IARC parameter’s development over time affects each growth function in a similar fashion: Imperfect learning of the NS parameter early on, corrected later, converging on the obligatory-subject target grammar. Based on the psycholinguistic data presented in \citet{10.2307/23358022}, one would expect to see an adjustment in the English-speaking child’s grammar away from an NS grammar, as children grow to interpret subjectless imperative sentences correctly as imperatives (as modeled by the IARC parameter). The simulations conducted in this study reflect this trajectory of the NS parameter, which supports the conjecture presented in \citet{10.2307/23358022} \textemdash{} that the misinterpretation of subjectless imperatives is indeed a likely contributor to a child’s Null Subject (NS) stage.

\bibliography{main}

\clearpage

\appendix
\section{Appendix}
\begin{figure*}[t]
\centering
\includegraphics[width=\textwidth]{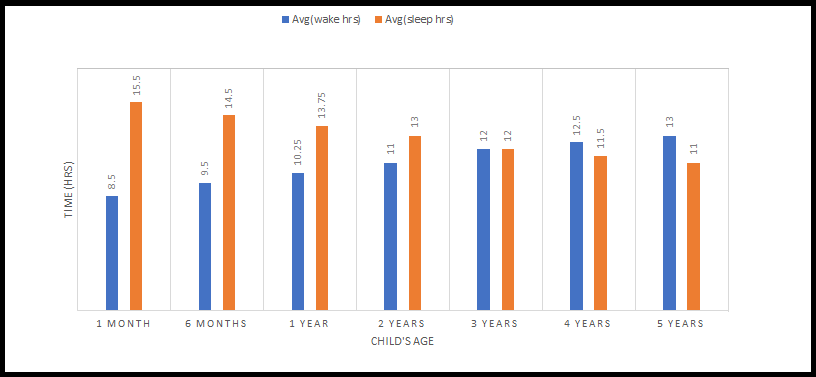}
\caption{Average total daily sleep and waking hours for infants and young children. Data is taken from \citet{DAVIS200465}.}
\label{fig4}
\end{figure*}
\subsection{Convergence\label{secconv}}

\begin{table*}[t]
\centering
\begin{tabular}{@{}c||cccccc@{}}
\toprule
\begin{tabular}[c]{@{}c@{}}age range \\($i$)\end{tabular}                             & \textbf{0;0 to 0;6} & \textbf{0;6 to 1;0} & \textbf{1;0 to 2;0} & \textbf{2;0 to 3;0} & \textbf{3;0 to 4;0} & \textbf{4;0 to 5;0} \\ \midrule
\begin{tabular}[c]{@{}c@{}}age period \\($a_i$)\end{tabular}                          & 0.5        & 0.5        & 1          & 1          & 1          & 1          \\ \midrule
\begin{tabular}[c]{@{}c@{}}daily waking hours\\ ($h^1_i $ to $h^2_i$)\end{tabular} & 8.5-9.5    & 9.5-10.25  & 10.25-11   & 11-12      & 12-12.5    & 12.5-13    \\ \midrule
\begin{tabular}[c]{@{}c@{}}total waking hours \\ ($H_i$)\end{tabular}              & 1,642.5    & 1,802.19   & 3,878.13   & 4,197.5    & 4,471.25   & 4,653.75   \\ \midrule
\begin{tabular}[c]{@{}c@{}}total utterances\\ ($u_i$)\end{tabular}               & 799,898    & 877,665    & 1,888,647  & 2,044,183  & 2,177,499  & 2,266,376  \\ \midrule
\begin{tabular}[c]{@{}c@{}}cumulative utterances\\ ($U_i$)\end{tabular}         & 799,898    & 1,677,563  & 3,566,210  & 5,610,392  & 7,787,891  & 10,054,267 \\ \bottomrule
\end{tabular}
\caption{Estimation of number of utterances encountered over different age ranges of child language acquisition.}
\label{table4}
\vspace{-4mm}
\end{table*}

Convergence is the learner\rq s arrival at a final grammar hypothesis ($G_{targ}$). The final grammar hypothesis should license nearly all utterances of the target language and generate the same set of sentences. Under standard learnability assumptions, convergence is defined as arriving at a static grammar, i.e., one that will never change within a finite amount of time after entertaining a series of grammar hypotheses --- \citet{GOLD1967447}, c.f., PAC-learning, \citet{Valiant:1984:TL:1968.1972}. 

Integrating finiteness into a criterion of success is desirable in terms of formal learnability theory, and from an empirical standpoint \textemdash{} developmental psycholinguistic studies have established a period during which language learning occurs rapidly and apparently effortlessly. After this \textit{critical period} \cite{penfield2014speech, lenneberg1967biological}, the learner achieves a state of maturity with less plasticity in terms of language development (i.e., the learner converges on an adult grammar). 

The implementation of this \textit{finiteness criterion} varies between studies. For example, in \citet{sakas2017parameter} the criterion of successful convergence for the variational learner was a parametric weight threshold of 0.02 from the target parameter setting for each parameter, and in the case that the threshold was not met, the simulations were stopped after an e-child encountered 2 million utterances. Whereas, for the No-Defaults Learner in \citet{doi:10.1080/10489223.2020.1803329}, simulations ended after an ad hoc number of sentences (500,000) were encountered by an e-child. 

\citet[Appendix A, Table 9]{doi:10.1080/10489223.2021.1888295}, estimate the number of sentences a real child hears between 2;4 and 5;0. They assume learning starts at 2;4 and calculated that from 28 months to 5 years a child from a professional family hears roughly 5,658,535 sentences. This calculation was based on \citet{hart1995meaningful,hart2003early}, who provide data on how many sentences professional class parents speak to their children and \citet{DAVIS200465} who provide the average total daily sleep hours for children.  In our case, however, we assume acquisition of the NS parameter starts at birth and estimate the number of sentences from birth to 5;0. We used  \citet[Figure 1]{DAVIS200465}, which plots daytime and nighttime sleeping hours to plot total waking and total sleeping hours by age, see Figure \ref{fig4}.


Using the data presented in Figure \ref{fig4}, we estimate the number of sentences a child hears from birth to age 5;0. In order to develop the relevant calculations, we adopted three assumptions:

\begin{enumerate}
    \item The number of waking hours of a child at age 1 month is almost the same as at birth.
    \item The number of utterances per hour spoken by a parent to a child is uniform across all ages, i.e., 487 \cite{hart1995meaningful,hart2003early}.\footnote {\citet{doi:10.1080/10489223.2021.1888295} make a similar assumption.}
    \item The increase in waking hours across age intervals is linear.
\end{enumerate}

When presenting our calculations, we employ the following notation. The \textit{age period} ($a_i$) is the difference in years, between two points delineating a specified \textit{age range} (\textit{i}). The \textit{daily waking hours} ($h^1_i $ to $h^2_i$) are the waking hours at the two endpoints of age range $i$. The \textit{total waking hours} of a child in age range $i$ is represented by $H_i$. \textit{Total utterances} ($u_i$) is the total number of utterances heard by the child in age range $i$ while the \textit{cumulative utterances} ($U_i$) is the total number of utterances heard by a child from birth to the last date of age range $i$.

We now turn to how we calculate some of these variables. To calculate total utterances in  age range $i$ ($u_i$), and subsequently cumulative utterances by the end of age range $i$ ($U_i$), we must first calculate the total waking hours at that age range ($H_i$). Figure \ref{fig4} gives the number of waking hours at specific ages. Assuming the growth of waking hours between any two adjacent ages is linear (Assumption 3) \textemdash{} to calculate the total waking hours between two adjacent ages, we compute the area under the straight ``line'' of growth between the two age intervals and multiply the area by the number of days in a year (365), see Equation (\ref{eq3}).

\begin{equation}
\label{eq3}
    H_i=\frac{(h^1_i+h^2_i)}{2}\times a_i \times 365
\end{equation}

The total utterances at age range $i$ ($u_i$) is then derived, under Assumption (2) by Equation (\ref{eq4}):
\begin{equation}
\label{eq4}
    u_i=\lceil H_i\times 487 \rceil
\end{equation}

Finally, we can then calculate the cumulative utterances at the end of age range $i$ ($U_i$) using Equation (\ref{eq5}):

\begin{equation}
\label{eq5}
    U_i= U_i+U_{i-1}
\end{equation} 

The results of these calculations are presented in Table \ref{table4}. Following \citet{doi:10.1080/10489223.2021.1888295}, we take the \textit{stopping point} for our simulated e-children to be 5;0. The number of cumulative utterances at 5;0 per our calculations is 10,054,267.

We can also approximate the number of cumulative utterances heard by a child at any given age. For example, to calculate the utterances heard by a child at age 3.3 years, we first need to approximate the waking hours at that age. The difference between the number of waking hours between ages 3;0 (12 waking hours) and 4;0 (12.5 waking hours) is 0.5 hours. Since we assume linear growth, we can approximate the number of waking hours at age 3.3 years: $12.15=12 + (0.3 * 0.5$). From Table \ref{table4}, we know that the number of cumulative utterances at age 3 years is 5,610,392. The total utterances a child hears between 3 years and 3.3 years can be calculated according to Equations (\ref{eq3}) and (\ref{eq4}), as is illustrated in (\ref{eq6}):
\begin{equation}
\begin{split}
1,322.2=\frac{12+12.15}{2}\times 0.3\times 365\\
643,918=\lceil1,322.2\times 487\rceil
\end{split}
\label{eq6}
\end{equation}

The cumulative utterances heard by age 3.3 years can then be calculated using Equation (\ref{eq5}): $6,254,310=643,918+5,610,392$.

\subsection{CoLAG domain details}

\begin{table*}[t]
\centering
\begin{tabular}{l l l l } 
 \hline
      & \multicolumn{2}{c}{Parameter List} \\[0.5ex] 
 \hline
 Parameter Name & Abbrev & Target Value = 0.0 & Target Value =1.0 \\ [0.5ex] 
 \hline\hline
Subject Position &(SP) & Initial & Final \\ 
Headedness in IP &(HIP) & Initial & Final  \\ 
Headedness in CP &(HCP) & Initial  & Final  \\ 
Optional Topic &(OpT) & Obligatory Topic & Optional Topic \\
Null Subject &(NS) & No Null Subject  & Optional Null Subject  \\
Null Topic &(NT) & No Null Topic  & Optional Null Topic  \\
Wh-Movement &(WhM) & Wh-Insitu  & Obligatory Wh Movement  \\
Preposition Stranding &(PI) & Obligatory Pied Piping & Optional Preposition Stranding \\
Topic Marking &(TM) & No Topic Marking & Obligatory Topic Marking \\
V to I Movemnt &(VtoI) & No VtoI Movement & Obligatory VtoI Movement \\
I to C Movement &(ItoC) & No ItoC Movement & Obligatory ItoC Movement \\
Affix Hopping &(AH) & No Affix Hopping & Affix Hopping \\
Question Inversion &(QInv) & No QInversion & Obligatory QInversion \\
 \hline
\end{tabular}
\caption{The 13 CoLAG parameters and their corresponding target values.} 
\label{tablecolag}
\vspace{-4mm}
\end{table*}

Thirteen syntactic parameters were used to generate the languages and derivations in CoLAG (see Table \ref{tablecolag}). The target parameter values of CoLAG English are: 0001001100011 which corresponds from left to right, the values of the thirteen parameters in Table \ref{tablecolag} from top to bottom. CoLAG English has word order patterns made up of the following lexical tokens: \textit{S}, \textit{01}, \textit{02}, \textit{03}, \textit{P}, \textit{Adv}, \textit{Aux}, \textit{Verb}, \textit{not}, and \textit{never}. These tokens correspond to subject, direct object, indirect object, object of a preposition, preposition, adverb, auxiliary, main verb, not and never respectively. CoLAG sentence patterns also have an overt (audible by e-children) illocutionary force feature: \textit{Q}, \textit{DEC} and \textit{IMP} for questions, declaratives and imperatives respectively. An example English pattern in CoLAG is: \textit{S Aux V O1 [DEC]} which might correspond to the natural language sentence: `\textit{The little dragon is breaking the wall.}'.


CoLAG English has 360 distinct sentence patterns, 180 declaratives, 36 imperatives, and 144 questions.  The Null Subject (NS) parameter is the parameter of interest here. If a CoLAG language is generated with NS=0 (e.g., CoLAG English), then every declarative and question has an overt subject. If NS=1, two versions of an utterance are generated, one with a subject and one without. The simulation studies detailed in this study present declaratives, questions, and imperatives to an e-child immersed in a CoLAG English-like language. Declaratives and questions are presented with overt subjects in CoLAG English. In CoLAG, imperative word orders universally do not have overt subjects.

\subsection{Additional Algorithms}

\begin{algorithm}[h!]
\small
	\SetAlgoLined
	\For{each $w_i$ in $W$}{
		set $w_i$ to 0.5. \;
	}	
	
	\For {each input sentence s}{
		\For {i in range(n) }
		    {
				with probability $w_i$, parameter value $p_i^v \gets 1$ \;
				with probability $1-w_i$, parameter value $p_i^v \gets 0$ \; 
			}
		$G_{curr}$ = [$p_1^v$, ... ,$p_n^v$]\;
		
		\uIf {$G_{curr}$ can parse s}{
		    \For {$w_i$ in $W$}{
		        adjust $w_i$ towards $p_i^v$ using Equation (\ref{VLreward0}) or (\ref{VLreward1});
		    }
		}
	}
\caption{Variational Learner reward only.}
\label{VL}
\end{algorithm}

\begin{algorithm}[h!]
\small
\SetAlgoLined
    IARC-list\textsubscript{1} $\gets$ sorted distribution of IARC for a 100 e-children of ages 2;6-2;11\\
    IARC-list\textsubscript{2} $\gets$ sorted distribution of IARC for a 100 e-children of ages 3;0-3;5\\
    IARC-list\textsubscript{3} $\gets$ sorted distribution of IARC for a 100 e-children of ages 3;6-3;11\\
    age-list\textsubscript{1} $\gets$ sorted distribution of ages for a 100 e-children of ages 2;6-2;11\\
    age-list\textsubscript{2} $\gets$ sorted distribution of ages for a 100 e-children of ages 3;0-3;5\\
    age-list\textsubscript{3} $\gets$ sorted distribution of ages for a 100 e-children of ages 3;6-3;11\\  
    \For {i in range (0 to 100)}{
        $IARC_1 \gets$ IARC-list\textsubscript{1}[i]\\
        $IARC_2 \gets$ IARC-ist\textsubscript{2}[i]\\
        $IARC_3 \gets$ IARC-list\textsubscript{3}[i]\\
        $age_1 \gets$ age-list\textsubscript{1}[i]\\
        $age_2 \gets$ age-list\textsubscript{2}[i]\\
        $age_3 \gets$ age-list\textsubscript{3}[i]\\
        Calculate optimal $m$ and $c$ using ($IARC_1$, $age_1$), ($IARC_2$, $age_2$), ($IARC_3$, $age_3$)\\
        Run Algorithm \ref{TVJ1} with optimal $m$ and $c$\\
    } 
 \caption{Simulating the TVJ experiment for a 100 e-children}
 \label{TVJ100}
\end{algorithm}
\end{document}